%% file: Learning_The_Network_of_Graphs_for_Graph_Neural_Networks.tex
\documentclass[journal]{IEEEtran}

\input{preamble.tex}
\usepackage{booktabs}
\newcommand{\relu}[1]{\mathrm{ReLU }\left(#1\right)}
\newcommand{\softmax}[1]{\mathrm{Softmax }\left(#1\right)}

\newcommand{\flatfun}[1]{\mathrm{Flat }\left(#1\right)}

\definecolor{green}{RGB}{51, 153, 0}

\ifCLASSINFOpdf
\else
\fi
\begin{document}
%
\title{ Learning the Network of Graphs for Graph Neural Networks}
%
%
%



\author{Yixiang~Shan$^\dagger$,~\IEEEmembership{Student~Member,~IEEE,}
	Jielong~Yang$^*$,~\IEEEmembership{Member,~IEEE,}
	Xing~Liu$^\dagger$,~\IEEEmembership{Member,~IEEE,}
	Yixing~Gao, Hechang~Chen, and~Shuzhi~Sam~Ge,~\IEEEmembership{Fellow,~IEEE}
\thanks{Yixiang Shan, Jielong Yang, Yixing Gao and Hechang Chen are with the School of Artificial Inteligence, Jilin University, Changchun, P.R. China (e-mail: shanyx20@mails.jlu.edu.cn; jyang022@e.ntu.edu.sg; gaoyixing.125@hotmail.com; chenhc@jlu.edu.cn).}
\thanks{Xing Liu is with the School of Astronautics, Northwestern Polytechnical University, Xi'an, P.R. China (e-mail: xingliu@nwpu.edu.cn).}
\thanks{Shuzhi Sam Ge is with the Robotics Research Lab, Department of Electrical \& Computer Engineering, National University of Singapore, Singapore (e-mail: samge@nus.edu.sg).}
\thanks{$^\dagger$Equal contribution. $^*$Corresponding author.}
}

%
%

\markboth{Journal of \LaTeX\ Class Files,~Vol.~14, No.~8, August~2015}%
{Shell \MakeLowercase{\textit{et al.}}: Bare Demo of IEEEtran.cls for IEEE Journals}
%



\maketitle

\begin{abstract}
	Graph neural networks (GNNs) have achieved great success in many scenarios with graph-structured data. However, in many real applications, there are three issues when applying GNNs: graphs are unknown, nodes have noisy features, and graphs contain noisy connections. 
	Aiming at solving these problems, we propose a new graph neural network named as GL-GNN. 
	Our model includes multiple sub-modules, each sub-module selects important data features and learn the corresponding key relation graph of data samples when graphs are unknown. GL-GNN further obtains the network of graphs by learning the network of sub-modules. The learned graphs are further fused using an aggregation method over the network of graphs.  
	Our model solves the first issue by simultaneously learning multiple relation graphs of data samples as well as a relation network of graphs, and solves the second and the third issue by selecting important data features as well as important data sample relations. We compare our method with 14 baseline methods on seven datasets when the graph is unknown and 11 baseline methods on two datasets when the graph is known. The results show that our method achieves better accuracies than the baseline methods and is capable of selecting important features and graph edges from the dataset. 
	Our code is publicly available at \url{https://github.com/Looomo/GL-GNN}.
\end{abstract}

\begin{IEEEkeywords}
graph neural network, learning discrete structures, network of graphs
\end{IEEEkeywords}

%
\IEEEpeerreviewmaketitle

\section{Introduction}
%
%
%
%
\label{Introduction} 
\IEEEPARstart{G}{raph} neural networks (GNNs) have achieved great success in many fields, such as social networks \cite{wu2019graph,qiu2018deepinf}, recommendation systems \cite{ying2018graph,xu2019graph}, pharmacology ~\cite{lin2020kgnn} and computer vision systems ~\cite{xu2017scene,landrieu2018large,cui2019dressing}. 
One of the advantages of GNNs is that it is able to exploit the relation information in the data. By incorporating the data relations, the learning algorithms can usually augment features with information from relations \cite{monner2013recurrent}
and thus achieve better performance. For example, in article topic identification problems, the citations among articles often represent similar themes, and thus can be used to identify article topics in a semi-supervised way. 
Semi-supervised learning methods are a set of algorithms that combine a small portion of labeled data with a large amount of unlabeled data during training ~\cite{liu2012robust}. GNNs are widely used to deal with semi-supervised  classification problems, in which each data sample is regarded as a node in the graph and the unlabeled nodes in the graph are predicted using the model trained by node features and training labels.

However, graphs are usually imperfect in real applications. 
In some cases, noisy node features and noisy node relations (i.e., graph edges) may present in a dataset. For example, in recommendation systems, both the social network information and the user information are noisy, and thus selecting the right social network connection and user information is critical to successful recommendations. The solution to these problems can greatly expand the applications of GNNs.
In some other cases, data does not contain a graph and the algorithm needs to jointly learn the graph structure and the model parameters. 
In these cases, some algorithms use two-stage procedures. These algorithms first use traditional methods to learn the graph, and then use GNN methods to solve the target problem. The most commonly used method
to learn a graph is $k$ nearest neighbor ($k$NN)\cite{altman1992introduction,ge2008hand}. NLE\cite{ge2010neighborhood} is another recently proposed intrinsic structures learning algorithm, which selects neighbors adaptively according to the inherent properties of samples. Inspired by NLE, DLLE\cite{ge2008hand} proposed a new neighbor finding method by estimating the probability density function of the input data.

For the previous methods, the graph construction and the model parameter learning are conducted separately, and the constructed graphs can not be adjusted in the training process. Thus, in some recently proposed methods, the graph structure and the model parameters are learned simultaneously.
In ~\cite{franceschi2019learning}, a method named LDS is proposed and can jointly learn the graph structure and the model parameters by solving a bilevel program. 
Graphite\cite{grover2019graphite} is a recently proposed unsupervised model used to learn the representations of nodes based on variational autoencoders, which uses GNNs to parameterize the encoders as well as decoders, and uses a multi-layer iterative process to build and refine the graph structure.


However, the relations among samples are due to many correlated factors. The previous methods consider neither the different factors nor the relations among these factors.

In this article, we propose a new graph neural network called GL-GNN, which is
capable of learning the network of graphs\footnote{In this paper, the network is also a graph whose nodes are graphs learned in sub-modules.} and using the obtained network to
learn graphs from different aspects. Our algorithm is composed of multiple sub-modules and each sub-module learns a relation graph of data samples. The network of graphs are constructed by learning a relation network of sub-modules and the aggregation module. The learned graphs are further fused with a aggregation method over the network of graphs. Each sub-module selects important node features and learns the corresponding relation graph of data samples from different feature aspects. These sub-modules have different parameters but the same neural network structure. Both relation graphs of data samples and the graph relation network are learned simultaneously. 

To the best of our knowledge, this is the first GNN that capable of learning the network of graphs, the important data features and the key relation graph of data samples simultaneously.

Our model can also provide interpretable explanations for predictions. We follow the definition of interpretable explaination in GNN Explainer proposed in \cite{ying2019gnnexplainer}. Specifically, the explanation of a predicted value $\hat{y}$ of the model is defined as ($G_S$, $X_S$), where G$_S$ is a subgraph of the data sample relation graph, $X_S$ is a subset of node features, and $G_S, X_S$ are the most relevant features and sub-graphs to the predicted values $\hat{y}$. We use the indicator vector to remove the noisy features and reconstruct the relation graph of data samples to remove noisy edges in the given graph. The remaining features and relations are thus the most relevant part to explain our predictions.


Our main contributions are as follows:
\begin{itemize}
	\item We propose a novel GNN that can use the network of graphs to simultaneously learn graph structures, the important data features, and the key relation graph of data samples. To the best of our knowledge, this is the first model that use the network of graphs to learn graph structures for GNNs.
	\item Our method simultaneously learns graphs and the semi-supervised classifier, and thus the graphs can be adjusted in the training process of the classifier. This property allows our method to be applied to the scenarios where graphs are unknown, nodes have noisy features, and graphs contain noisy connections, which expand the application of GNNs. 
	
	\item We compare our method with 14 baseline methods when graph is not available and 11 baseline methods when graph is available, including latest methods capable of learning graphs and two-stage methods that use kNN to learn graphs. The results show that our method achieves better performance than the baseline methods.
	\item 
	Experiments on 7 real datasets show our algorithm can remove the noisy subset of features and noisy relations of nodes (i.e., learn the key features and key graph edges).
\end{itemize}

The rest of the paper is structured as follows. \cref{Related Works} presents the related work. \cref{Approach} illustrates the problem and our proposed algorithm. \cref{Expirements} evaluates the performance of our method and 24 baseline methods in both cases where the graph is available and unavailable. \cref{conclusion} draws the conclusions. 

\emph{Notations:} We use boldfaced characters to represent vectors and matrices. Suppose that $\bM$ is a matrix, then $\bM_{m,\cdot}$, $\bM_{\cdot,m}$, and $\bM_{m,n}$ denote its $m$-th row, $n$-th column, and $(m,n)$-th element, respectively. The vector $(x_1,\ldots,x_N)$ is abbreviated as $(x_n)_{n=1}^N$. Let $\bI$ be the identity matrix. Let $\softmax{\cdot}$ be the softmax function. Let $\text{Concat}(\cdot)$ represents the concatenate operation, $\flatfun{\cdot}$ represents the flatten operation.

\section{Related Works}
\label{Related Works}
\subsection{Graph Neural Networks}
In \cite{defferrard2016convolutional}, Graph-CNN is proposed by computing the eigenvalues of the normalized Laplacian matrix of the graph. The Chebyshev polynomials of the diagonal matrix of these eigenvalues are then employed as a filter for the subsequent graph convolution. In \cite{kipf2019semi}, GCN is proposed and it uses the first-order approximation of Graph-CNN, which achieves better performance in semi-supervised learning problems. The forward model of GCN is given by
\begin{align}
	Z=\text{Softmax}(\widehat{\boldsymbol{A}}\text{ReLU}(\widehat{\boldsymbol{A}}\boldsymbol{X}\boldsymbol{W}^{(0)})\boldsymbol{W}^{(1)}),
\end{align}
where $\widehat{\boldsymbol{A}}$ is the normalized adjacency matrix, $\boldsymbol{X}$ is the input data and $\boldsymbol{W}^{(0)}$ and $\boldsymbol{W}^{(1)}$ are two trainable matrices in layer 0 and layer 1.
In \cite{velivckovic2018graph}, GAT is proposed and it learns the attention coefficient between each pair of linked nodes and use these coefficients to aggregate messages for each node from its neighbors.

\subsection{Graph Learning for GNNs}
In order to use graph neural network methods in applications without graph, we need to learn graphs from the data. kNN is a simple but efficient method to learn graph structures and thus is widely used in two-stage methods that first learn graphs and then use fixed graphs to learn graph neural networks.

In \cite{franceschi2019learning}, LDS uses the bilevel optimization algorithm to simultaneously learn the discrete and sparse dependencies among data samples and the parameters of the graph convolutional network (GCN). LDS is the first method to simultaneously learn graphs and neural network  parameters in a semi-supervised node classification problem and achieves better performances than the two-stage methods. This route are then followed by recently proposed DIAL-GNN \cite{chen2019deep}. 
Our proposed method also follows this route but introduced a network-of-graphs based approach to learn the graph and neural network parameters.

%

\subsection{Link Prediction}
The work similar to the graph learning is link prediction. Most link predictions are in the form of  similarity measure, but there are also methods based on deep learning, such as Wlnm\cite{zhang2017weisfeiler}. Some algorithms use probabilistic generative models to learn the graph structure. The probabilistic generation model is originated from random graph generation model\cite{erdHos1960evolution}, and later derived many other algorithms such as degree distribution \cite{leskovec2005graphs}. 

\subsection{Prediction Explanation}
Most deep learning models are black boxes, which prevents developers from selecting useful data information and results in inefficient usage of computing resources\cite{ribeiro2016should}. Thus, prediction explanation of deep neural networks is an important task.  
In \cite{ribeiro2016should}, the authors propose an approach called LIME which can provide faithful interpretability for many classifiers by approximating the original model using another interpretable  model. A method called SP-LIME is also proposed in \cite{ribeiro2016should} and the method provides an intuitive global model explanation by selecting a representative set with explanations. However, these algorithms are not designed for GNNs. Thus, in \cite{ying2019gnnexplainer}, Ying $et~al.$ propose GNNExplainer, which generates a consistent and concise explanation for the GNN model by learning an important sub-graph and an important subset of node features.

%
\begin{figure*}
	\centering

	\includegraphics[width=0.85\linewidth]{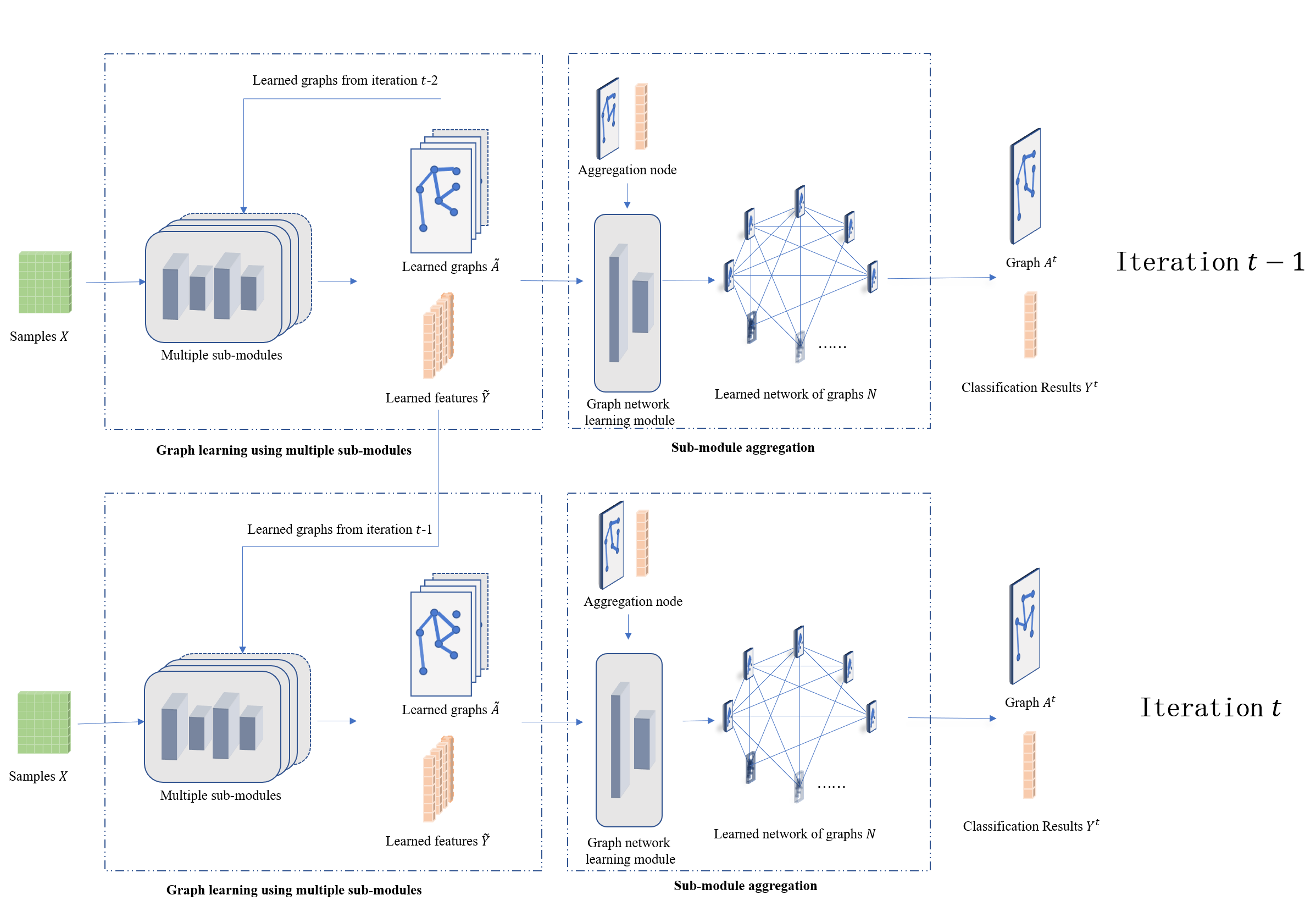}
	\caption{The schematic representation of our algorithm. Our model includes two parts: a graph learning algorithm with multiple sub-modules and a graph network learning algorithm. Our method learns $m$ graphs of samples in $m$ sub-modules and in the graph network learning algorithm, where a network of sub-modules is learned and then used to fuse the sub-module results. The schematic representation of each sub-module is shown in \cref{fig:sub-module}. 
	} 
	
	\label{fig:structure}	
	
\end{figure*}
\section{Approach}
\label{Approach}
We consider $N$ data samples, $\bX=\{\bx_i\}_{i=1}^N$, where $\bx_i$ $\in$ $\boldsymbol{R}^F$ and $F$ is the dimension of feature. The one-hot labels of the data samples are $\bY=\{\by_i\}_{i=1}^N$, where $\by_i \in R^{C}$ and $C$ is the number of classes. In many machine learning methods, sample independence is an underlying assumption. However, in many practical applications, there are intrinsic relations among data samples. Taking advantage of these relations will improve the algorithm performance. 

Traditional graph neural network methods have achieved good performance in semi-supervised classification tasks on graph-structured datasets. However, the graph is often noisy or completely unavailable. Hence, in this paper, we aim at applying our GNN to the scenarios when graphs are not available, nodes have noisy features and graphs contain noisy connections. Specifically, we conduct semi-supervised node classification in considering of the following situations: 1) When the graph is unknown, we learn the relations among nodes (i.e., graphs of data samples) and remove the unimportant features of the data samples; 2) When the graph is known, we remove the unimportant features and the noisy edges in the graph.

\subsection{Overall Model}
We define a graph as a triple $\{\calV,\calE, \calW \}$, where $\calV = \{1,2,....,N\}$ is a node set, $\calE \subseteq \calV \times \calV$ is an edge set, and  $\calW:\calE\rightarrow R$  is a mapping from edge set to $R$, representing the weights of edges.

\begin{figure*}	
	\centering
	\includegraphics[width=0.85\linewidth]{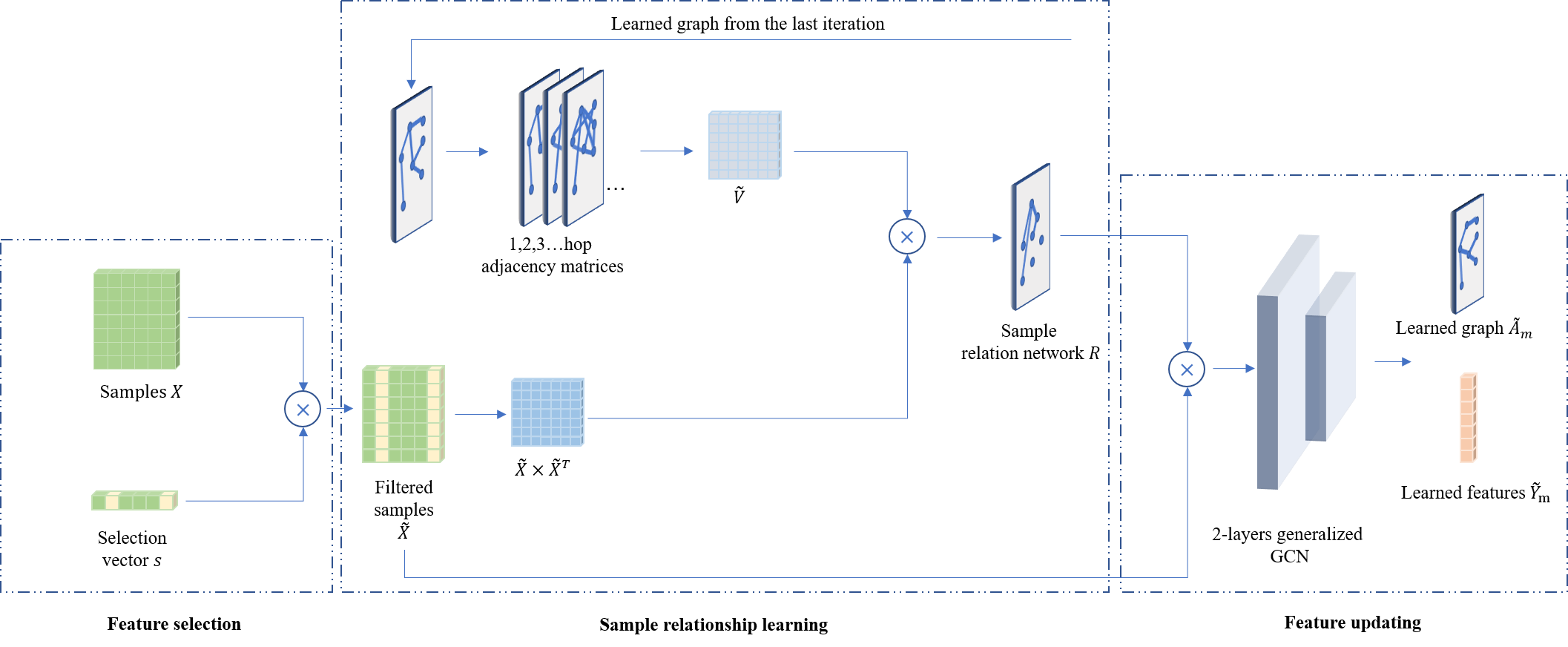}	
	\caption{The schematic representation of each sub-module. Each sub-module contains three parts: feature selection, sample realtionship learning and feature updating. In the feature selection part, features marked with yellow blocks in the selection vector are those to be filtered out. In the sample relation learning algorithm, each sub-module learns the relation graph of data samples with connection indices sampled from a discrete distribution, which is discussed in \cref{graph_learning}. $[\cdot]^{T}$ represents the transposition operation.
}

	\label{fig:sub-module}

\end{figure*}

The overall structure of our method is shown in \cref{fig:structure}. 
Data samples are input into $M$ sub-modules to learn $M$ sample relation graphs independently.
$M$ graphs as well as $M$ learned features are further input into the graph network learning module to learn a network of graphs, which is then used in graph aggregation module to generate graphs and classification results. The overall structure of each sub-module is shown in \cref{fig:sub-module}.

The parameters and graph structures are learned simultaneously and thus the discrete structure we learned can be adaptively adjusted in the neural network learning process. Our model includes two parts: a graph learning algorithm using multiple sub-modules and a network of graphs learning algorithm.  In the graph learning algorithm, each sub-module selects the node features and learn the corresponding relation graph of data samples. In the graph network learning algorithm, we learn the network of sub-modules. 

The first part of our model, a graph learning algorithm using multiple sub-modules is illustrated in \cref{fig:sub-module}. The second part of our model, the network of Graphs learning algorithm is illustrated in \cref{fig:structure}.

\subsection{Graph Learning Using Multiple Sub-modules}\label{sec:Graph_learning}

Each sub-module includes three parts: key feature learning, key graph learning and feature updating.

\subsubsection{Key Feature Learning} 
\label{Feature selection}

The noisy features can make the model learning difficult\cite{sung2003identifying}. In this paper, we use multiple selection vectors $(\bs_m)_{m=1}^M$ to select different subsets of data features. 
Different sub-modules in our model share the same structure but have independent parameters.
We take sub-module $m$ as an example in this section. Sub-module $m$ has a feature selection vector $\boldsymbol{s}_m\in R^F$.  We select the features by multiplying each row of $\boldsymbol{X}$ with $\boldsymbol{s}_m$ and obtain the filtered features

\begin{equation}
	\tbX=(\tbx_{m,i})_{m=1,i=1}^{M,N},\label{selectfeature}
\end{equation}
where $\tbx_{m,i}=\boldsymbol{x}_i\odot\boldsymbol{s}_m$ and $\odot$ represents element-wise multiplication. We will show in \cref{sec45} that selection vectors $\boldsymbol{s}$ also provide an interpretable explanation of the predictions.  

We initialize the selection vectors $(\boldsymbol{s}_m)_{m=1}^M$ randomly at the beginning, different $\boldsymbol{s}_m$ will have different values and thus different sub-modules focus on different subsets of features. Different selection vectors will further learn different data sample relation graphs, thus encourage the $(\boldsymbol{s}_m)_{m=1}^M$ to select different features in the next iteration. At the end of the learning process, the smaller values of $\boldsymbol{s}_m$ will indicate the noisy dimensions of features.  We then aggregate sub-modules in the sub-module network learning part, which will be introduced in \cref{Sub-module relation learning}.

\subsubsection{Key Graph Learning} \label{sec311}
\label{graph_learning}
We use $\tbX$ to learn the data sample relation graphs. 

In our method, we treat labels $\boldsymbol{y}$ of the training data samples as a feature and concatenate it with $\tbX$ when learning the graph. For training data sample $i$, we obtain the concatenated features
\begin{align}
	\boldsymbol{Z}_{m,i} = \text{Concat}	\left(\boldsymbol{y}_i,\tbX_{m,i}\right)\label{eq:concatenate}.
\end{align}
When evaluating our method on test datasets, we replace the concatenated labels with zeros.

Apart from learning the relations among samples (i.e., the data sample graph) in each iteration, our model also exploits the graphs learned in the last iteration.


We use discrete random variables $\bW_{i,j}=1$ ($0$) to denote sample $i$ is (not) connected to sample $j$, and use the following discrete distribution to model $\bW_{i,j}$:
\begin{equation}
	\begin{aligned}
		&\bP(\bW_{i,j}=1|\bV_{i,j},\bZ_{m,i} , \bZ_{m,j})\\
		&=\bV_{i,j}\bP(\bW_{i,j}=1|\bZ_{m,i},\bZ_{m,j})\\
		&=\bV_{i,j}\text{Softmax}(\bZ_{m,i}^T\bZ_{m,j})  ,\label{eq:catdis}
	\end{aligned}
\end{equation}

where $\bZ_{m,i}, \bZ_{m,j}$ are the features of samples $i$ and $j$, which are obtained from \cref{eq:concatenate}. $\bV$ is a $N\times N$ variable, and $\bV_{i,j}$ represents the probability of the connection between sample $i$ and $j$. 

There are two difficulties in directly learning $\bW_{i,j}$ using \cref{eq:catdis}:
\begin{itemize}
	\item [$i$.] 
	
	$\bW_{i,j}$ is a discrete random variable. We cannot backpropagate gradients through $\bW_{i,j}$ \cite{kingma2013auto,yoon2020data} if we directly learn $\bW_{i,j}$.\label{diff2}
	
	\item [$ii$.]
	$\bV_{i,j}$ is a matrix with $N \times N$ elements. Too many parameters pose challenges for effective learning. \label{diff1}

\end{itemize}

For difficulty $i$, there are multiple ways to handle the non-differentiable problem, such as subgradient method and reparameterization (Gumbel-softmax\cite{jang2016categorical}, stochastic back-propagation\cite{kingma2014stochastic}). In this paper, we learn the parameter $\widetilde{\bV}_{i,j}$ of categorical distribution instead of directly learning $\bW_{i,j}$. 

For difficulty $ii$, 
considering that the influences between two nodes are related to their topological distances, for any $i,j,j'\in\{1,\ldots,N\}$, we let $\bV_{i,j}=\bV_{i,j'}=\widetilde{\bV}_{k}$ if node $j$ and node $j'$ are both the $k<K$ hop neighbors of node $i$, where $K$ is a known hyperparameter. For any two $k>K$ neighbors $j,j'$, we let $\bV_{i,j}=\bV_{i,j'}=\widetilde{\bV}_{o}$. This strategy reduces the complexity of optimizing $\bV_{i,j}$ from $O\left(N^{2}\right)  $ to $O\left(1\right)$ and thus speeds up the training process.

Specifically, we first use the following formula to calculate the multi-hop neighbor information of each sample based on the relation graph of the last iteration:

\begin{equation}
	\bH_{m,i,j}^{\lambda} = [\bA_{m,i,j}^{l}]^{\lambda},\label{eq:hop}
\end{equation}

where $\bA_{m,i,j}^{l}$ is the relation graph of the last iteration and is computed in \cref{eq:adj_sub}.

Combining \cref{eq:catdis} and \cref{eq:hop}, we have the sample relation graph in iteration $t$:
\begin{equation}
	\bR_{m,i,j} = \text{Softmax}(\bZ_{m,i}^T\bZ_{m,j})\cdot(\sum_{k = 1}^{K}\bH_{m,i,j}^{k} \widetilde{\bV}_{k} + \widetilde{\bH}\widetilde{\bV}_{o}), \label{adj_l1}
\end{equation}
where $\widetilde{\bH}$ represents the $k>K$ hop adjacency matrix.






Then we keep the top $k$ values of each row of $\boldsymbol{R}_{m,\cdot,\cdot}$ and set the other values of each row to 0 to obtain the adjacency matrix $\boldsymbol{A}_{m,i,j}$.

 The graph learned in our algorithm is changed with the learnable selection vector $\boldsymbol{s}_m$. We use $\boldsymbol{s}_m$ to select important features and then learn the corresponding graph of the selected features. 

The learned graph is then used to update the features.

\subsubsection{Feature Updating} 
\label{feature update section}
In this section, we use a 2-layers generalized GCN method to update features. Specifically, we first normalize the obtained adjacency matrix $\bA^{'}$:
\begin{align} 
	\boldsymbol{A}_{m,\cdot,\cdot}^{'}=\boldsymbol{D}^{-1/2}(\boldsymbol{A}_{m,\cdot,\cdot}+\boldsymbol{I})\bD^{-1/2},\label{eq:nor_adj}
\end{align} 
where $\bD$ is a diagonal matrix with $\bD(i,i)=1+\sum_j\bA_{m,i,j}$ for any $m\in\{1,...,M\}$.

The original output feature $\tbX$ is then updated as
\begin{align}
	\tbX_{m,i}^{(1)}&=\relu{\bA^{'}_{m,\cdot,\cdot}\tbX_{m,i} \bW^{(1)}},\label{layer1}\\ 
	\tbX_{m,i}^{(2)}&=\softmax{\bA^{'}_{m,\cdot,\cdot}\tbX_{m,i}^{(1)} \bW^{(2)}},\label{layer2}
\end{align}

where $\tbX_{m,i}^{(1)}$ and $\tbX_{m,i}^{(2)}$ represents the updated features in the first and second hidden layers of the generalized GCN method. $\relu{\cdot}$ represents the relu activation function, $\softmax{\cdot}$ represents the softmax activation function and $ \bW^{(1)},\bW^{(2)} $ represents the learnable weights in the first and the second hidden layers. 

We then compute another graph using $\tbX_{m,i}^{(1)}$:
\begin{align}
	\bZ^{(1)}_{m,i} &= \text{Concat}	\left(\by_i,\tbX^{(1)}_{m,i}\right),\label{concat_l2}
\end{align}
\begin{align}
	\bR^{(1)}_{m,i,j} &= \bZ^{(1)}_{m,i} \bZ^{T {(1)}}_{m,j}\label{adj_2}.
\end{align}
We also normalize the obtained adjacency matrix $\bR^{(1)}_{m,i,j}$

\begin{align}
	\boldsymbol{A}_{m,\cdot,\cdot}^{(1)'}={{\bD}^{(1)}}^{-1/2}(\boldsymbol{A}^{(1)}_{m,\cdot,\cdot}+\boldsymbol{I}){\bD^{(1)}}^{-1/2},
	\label{eq:adj_sub}
\end{align}

where $\bD^{(1)}$ is a diagonal matrix with $\bD^{(1)}(i,i)=1+\sum_j\bA^{(1)}_{m,i,j}$ for any $m\in\{1,...,M\}$. The graph $\bA^{(1)'}_{m,i,j}$ will not be used to update features, but will be used to calculate loss in \cref{loss_1} to ensure the learned graphs stay similar in the feature updating process.

Different from the original GCN method that computes the matrix $\bA_{m,\cdot,\cdot}^{'}$ before the training process, in our method $\bA_{m,\cdot,\cdot}^{'}$ is learned together with $\bW^{(1)}$ and $\bW^{(2)}$.


\subsection{Aggregation Using Network of Graphs}

\label{Sub-module relation learning}
GL-GNN learns a network of graphs by learning the relations among multiple sub-modules and use this network to aggregate the learned graphs and classification results.
In our graph learning algorithm, we use multiple sub-modules to filter features, learn the relation graphs of data samples and update filtered features.
Our graph learning algorithm focuses on different dimensions of features in different sub-modules and thus learns the data sample relation graphs from different aspects by applying a sub-module relation learning method. In this paper, we consider $M$ sub-modules.

We use a aggregation module to aggregate different sub-modules. In each iteration, we learn a relation network whose nodes are the aggregation module and the $M$ sub-modules. The learned relation network  is then used to aggregate the results of all the sub-modules. 

\subsubsection{Graph Network Learning}
\label{sec:graph_n_l}

The graph network learning algorithm plays an important role in our method. It can not only aggregate the training results of multiple sub-modules, but also adaptively assign weights to $M$ sub-modules. 

We construct the relation network of the aggregation module and $M$ sub-modules as an $(m+1)$ $\times$ $(m+1)$ matrix, and the attention coefficient of sub-module $m$ to $m'$ is given by
\begin{equation}
	\balpha_{m,m'}=\ba^T(  \flatfun{\tbX_{m,\cdot}^{(2)}} \odot  \flatfun{\tbX_{m',\cdot}^{(2)}} ),
\end{equation}

where $\balpha$ represents the attention coefficient matrix and $\ba$ is a learnable vector with the same shape as  $\flatfun{\tbX_{m,\cdot}^{(2)}}$. 
The attention coefficient of aggregation module $\bX_{agg}$ to sub-module $m$ is
\begin{equation}
\balpha_{m,agg}=\ba^T(    \flatfun{\tbX_{m,\cdot}^{(2)}} \odot \flatfun{\bX_{agg}} ). \label{attention}
\end{equation}

We also calculate the self attention coefficient of the aggregation node $\bX_{agg}$ :
\begin{equation}
\balpha_{agg,agg}=\ba^T(  \flatfun{\bX_{agg}} \odot  \flatfun{\bX_{agg}} ).
\end{equation}

To make the attention more flexible, we multiply $\balpha$ by another trainable variable $\bb$ and the result is given by:
\begin{equation}
\bN=\balpha\bb,
\end{equation}
where $\bN$ is regarded as the adjacency matrix of the network of graphs.




\subsubsection{Aggregation}
\label{nonlinear}

We use the network of graphs learned in \cref{sec:graph_n_l} to aggregate learned graphs and learned features. The adjacency matrix of the network of graphs is an $(m+1)\times (m+1)$ matrix $\bN$, in which $\bN_{m,m'}$ represents the attention coefficient of graph $m$ to $m'$. Let \text{Agg} and Agg$'$ be the methods for aggregating learned graphs and learned features, respectively. The scheme can be formulated as:

	\begin{align}
	\bA^t=\text{Agg}(\widetilde{\bA},\bN),\\
	\bY^t=\text{Agg}'(\widetilde{\bY},\bN),
	\end{align}
\label{eq:aggre}
where $\widetilde{\bA}=\{\widetilde{\bA}_{m}\}_{m=1}^M$ and $\widetilde{\bA}_{m}$ represents the $m$-th graph, $\bA^t$ is the aggregation result of graphs, $\widetilde{\bY}=\{\widetilde{\bY}_{m}\}_{m=1}^M$, $\widetilde{\bY}_{m}$ represents the $m$-th learned features, which is equal to $\tbX_{m,\cdot}^{(2)}$ in our method and $\bY^t$ is the class predictions of our model. 


Specifically, the the class predictions of our model $\bY^{t}$ in current iteration is

\begin{equation}
	 \bY^t=\softmax{\sum_{m=1}^{M}\bN_{agg,m}\tbX_{m,\cdot}^{(2)} +	\bN_{agg,agg}\bX_{agg}   },\label{fusing}
\end{equation}
and the graph $\bA^{t}$ is aggregated as

\begin{equation}
	\bA^t=\softmax{\sum_{m=1}^{M}\bN_{agg,m}\widetilde{\bA}_{m}  }.\label{fusing_adj}
\end{equation}

\subsection{Loss Function}
To ensure the graph in our feature updating process in  \cref{feature update section} stays similar, we define the following loss function:
\begin{align}
	L_1=\sum_{m=1}^M| \boldsymbol{A}^{'}_m -\boldsymbol{A}_m^{(1)'}|,\label{loss_1}
\end{align} 
where $\boldsymbol{A}^{'}_m$ is the adjacency matrix obtained in  \cref{eq:nor_adj} and $\boldsymbol{A}_m^{(1)'}$ is the adjacency matrix obtained in \cref{adj_2}. The loss $L_1$ is used to ensure the learned graphs in different layers of our model stay similar in each iteration.

Like many other graph neural network methods, our method is mainly used to deal with the node classification problem. We use the same loss function as in ~\cite{scarselli2008graph} given by
\begin{align}
	L_2=-\sum_{i=1}^N\sum_{c=1}^C \by_{i,c}\log \bg_{i,c},\label{loss_2}
\end{align}
where $C$ is the number of classes.
We sum up $L_1$ and $L_2$ and obtain the loss function $L$ used to train our model
\begin{equation}
	L=\mu_{L1}L_1+\mu_{L2}L_2 + L_{R}.\label{loss_3}
\end{equation}
Where $\mu_{L1}$ and $\mu_{L2}$ are the weights of loss  $L_1$ and $L_2$, and $L_R$ is the loss of regularization operation.

\subsection{The Pseudo Code of GL-GNN}
The pseudo code of GL-GNN is shown in \cref{{alg:GL-GNN}}. 

Line 4 to line 11 is corresponding to the graph learning algorithm using multiple sub-modules illustrated in \cref{sec:Graph_learning} and line 12 to line 14 is corresponding to the sub-module relation learning algorithm in \cref{Sub-module relation learning}. The stopping conditions of our algorithm means the iteration reaches the maximum number of the iterations.

\begin{algorithm}
	\caption{GL-GNN}
	\label{alg:GL-GNN}
	\begin{algorithmic}[1]
		\STATE \textbf{Input: Data samples $\bX$, labels of data samples $\bY$}
		\STATE \textbf{Variables:  $(\bs_m)_{m=1}^M$,  $\bb$, $\ba$, variables in  two generalized GCN layers}                       
		\WHILE{ stopping conditions are not satisfied} 
		\FOR{each sub-module $m\in\{1,\ldots,M\}$} 
		\STATE Select features using  $\tbx_{m,i}=\bx_i\odot\bs_m$.
		\STATE Learn a graph from selected features using formula \cref{eq:concatenate} and \cref{adj_l1}.
		
		\STATE Update features using a generalized GCN layer \cref{layer1}.
		
		\STATE Learn a graph from the updated features using formula \cref{concat_l2} and \cref{adj_2}.
		
		\STATE Update features using another generalized GCN layer \cref{layer2}.
		
		\ENDFOR
		\STATE Learn network of graphs using \cref{attention}.
		
		\STATE Aggregate the results of the sub-modules using \cref{fusing}
		\STATE Compute the loss using \cref{loss_1}, \cref{loss_2}, and \cref{loss_3}.
		\ENDWHILE
		
		\RETURN Learned variables in our GL-GNN model
	\end{algorithmic}
\end{algorithm}

\begin{table*}[t]
		\caption{Node classification accuracy when the graph is unknown. The best two  baseline methods have been marked in bold. } 
		\label{tab:acc without graph}
		\begin{center}
			
		\begin{tabular}{@{}lrrrrrrr@{}}
			\toprule
			Method	   & Wine          & Cancer        & Digits         & Citeseer       & Cora           & 20news                & FMA                            \\ \midrule
			LogReg     & 92.1          & 93.3          & 85.5           & 62.2           & 60.8           & 42.7                  & 37.3                           \\
			LinearSVM  & 93.9          & 90.6          & 87.1           & 58.3           & 58.9           & 40.3                  & 35.7                           \\
			RBF SVM    & \textbf{94.1} & 91.7          & 86.9           & 60.2           & 59.7           & 41.0                    & \textbf{38.3}                      \\ \midrule
			RF         & 93.7          & 92.1          & 83.1           & 60.7           & 58.7           & 40.0                    & 37.9                           \\
			FFNN       & 89.7          & 92.9          & 36.3           & 56.7           & 56.1           & 38.6                  & 33.2                           \\
			LP         & 89.8          & 76.6          & 91.9           & 23.2           & 37.8           & 35.3                  & 14.1                           \\
			ManiReg    & 90.5          & 81.8          & 83.9           & 67.7           & 62.3           & 46.6                  & 34.2                           \\
			SemiEmb\cite{weston2012deep}    & 91.9          & 89.7          & 90.9           & 68.1           & 63.1           & \textbf{46.9}         & 34.1                           \\
			Sparse-GCN\cite{erdHos1960evolution} & 63.5          & 72.5          & 13.4           & 33.1           & 30.6           & 24.7                  & 23.4                           \\
			Dense-GCN  & 90.6          & 90.5          & 35.6           & 58.4           & 59.1           & 40.1                  & \textbf{34.5}                           \\
			RBF-GCN    & 90.6          & 92.6          & 70.8           & 58.1           & 57.1           & 39.3                  & 33.7                           \\
			$k$NN-GCN\cite{franceschi2019learning}    & 93.2          & \textbf{93.8} & 91.3           & 68.3           & \textbf{66.5}  & 41.3                  & 37.8                           \\
			$k$NN-GAT    & \textbf{93.6}          & 74.5          & \textbf{92.8}  & \textbf{69.3}  & 65.7           & 41.3                  & 23.9                               \\
			$k$NN-LDS\cite{franceschi2019learning}    & \textbf{97.3} & \textbf{94.4} & \textbf{92.5}  & \textbf{71.5}  & \textbf{71.5}  & \textbf{46.4}         & \textbf{39.7}                  \\
			\midrule
			GL-GNN	   & \textbf{98.3} & \textbf{96.8} & \textbf{93.7} & \textbf{73.5}    & \textbf{74.0}  & \textbf{51.1}         & \textbf{40.0}                    \\
			\bottomrule
	\end{tabular}
	\end{center}

\end{table*}

\begin{table*}[]
	\caption{ Accuracy of GL-GNN on different datasets with or without noisy edges and the denoising performance of GL-GNN. The row "Remaining / Added Noisy Edges" represents the ratio of remaining noisy edges to the added noisy edges. Our algorithm can remove most of the noisy edges on all but Wine dataset. }
	\label{tab:denoise}
	\begin{center}
		
	\begin{tabular}{lrrrrrrr}
		\toprule
		& Wine      & Digits      & Cancer & Cora    & Citeseer & 20news & FMA    \\
		\midrule
		Accuracy without noisy edges      & 98.3 		& 93.6       & 96.8          & 85.6   & 75.3       & 50.1     & 40.0     \\
		\midrule
		Accuracy with noisy edges & 97.5      & 92.4        & 96.3          & 84.5    & 74.4     & 47.3     & 39.5  \\
		\midrule
		Remaining / Added Noisy Edges  & 3560/6390 & 3399/645397 & 5060/32475    & 24/7291 & 31/11031 & 38/46407 & 9/6397  \\
		\bottomrule
	\end{tabular}
	\end{center}

\end{table*}

\begin{table}[]
	\caption{Node classification accuracy when the graph is known.  The best two baseline methods have been marked in bold.   } 
	\label{tab:acc using graph}
	\begin{center}
		
	\begin{tabular}{@{}lrr@{}}
		\toprule
		Method       & Citeseer           & Cora          \\
		\midrule
		GCN(Semi Supervised)\cite{franceschi2019learning} & 70.3           & 81.5          \\
		SemiEmb\cite{weston2012deep}				& 59.6				& 59.0		\\
		DeepWalk\cite{perozzi2014deepwalk}				& 43.2				& 67.2		\\
		ICA\cite{getoor2005link}				& 69.1				& 75.1		\\
		Planetoid\cite{yang2016revisiting}				& 64.7				& 75.7		\\
		GMNN\cite{qu2019gmnn}                   & 73.0             & 83.4          \\
		GRCN\cite{yu2019graph}                   & \textbf{73.6}           & 84.2          \\
		GraphHeat\cite{xu2019graph}              & 72.5           & 83.7          \\
		GLCN\cite{jiang2019semi}                   & 72.0             & \textbf{85.5}          \\
		DIAL-GNN\cite{chen2019deep}  			   & \textbf{74.1} & \textbf{84.5}  \\
		Graphite\cite{grover2019graphite}  			   & 71.0			 & 82.1  \\
		\midrule
		GL-GNN                 & \textbf{76.3}  & \textbf{85.5}	\\
		\bottomrule
	\end{tabular}
\end{center}

\end{table}

\subsection{Time and Memory Complexity}
The overall time complexity and memory complexity of our method are $O( MT(N^2+N) )$ and $O( M(N^2+N) )$, where M, T, N are the numbers of sub-modules, epochs and nodes. In comparison, the time complexities of DIAL-gcn and LDS are $O(T(N^2+N))$ and $O(T_o N^2 T_i SC)$, where $T_o$ and $T_i$ are the numbers of the outer and the inner iterations, S is the number of graph samples and C is the number of edges.

\section{Experiments}
\label{Expirements}
We conducted 10 Monte-Carlo experiments on seven datasets when graph is not available as well as two datasets when graph is available and take the average of the results in this paper.
We compared GL-GNN with 14 baseline methods when the graph is unknown and 11 baseline methods when the graph is known. The experiment results are shown in \cref{tab:acc without graph}, \cref{tab:acc using graph} and \cref{tab:denoise}. GL-GNN achieves better performance than the baseline methods in node classification problem and shows good performance in important feature selection and important graph edge extraction.

\subsection{Datasets}
We use the following datasets: Wine, Cancer, Digits, Citeseer, Cora, 20news, and FMA. The information of these datasets are summarized in \cref{tab:my-table}. 
For the case where there is no external graph,we use the same data split settings as LDS method proposed in \cite{franceschi2019learning}. For the case where there is an external graph, we use the same data split settings as DIAL-GNN proposed in \cite{chen2019deep}.

Details of datasets used in this paper are as follows.

\subsubsection{Wine}

The Wine dataset contains 178 data samples. Each data sample has 13 features representing 13 chemical components of wines and the data samples belong to 3 classes representing three different origins of wines. Wine dataset does not contain daat sample graph, and is widely used in classification tasks.

\subsubsection{Cancer}

The Cancer dataset contains 569 samples, and each sample contains 30 features, which are extracted from photos of breast lumps fine needle aspiration (FNA). Each sample can be either benign or malignant. Cancer dataset includs 357 benign records and 212 malignant records but does not contain any graph.

\subsubsection{Digits}

The data samples in this dataset are images of hand-write digits. Each image is an $8\times 8$ pixel matrix. This dataset also does not contain any graph.

\subsubsection{Citeseer and Cora}
Cora and Citeseer are two commonly used benchmarks for evaluating graph-based learning algorithms. These two datasets include the keyword information and the citation networks of articles. In our algorithm, we use Cora and Citeseer datasets to evaluate our method in both the case where there is a graph and the case where there is no graph. 
\subsubsection{20news}

The 20 Newsgroup dataset is composed of 18000 newsgroups posts on 20 topics. Same as \cite{franceschi2019learning}, we select 10 classes from 20 Newsgroup and regard features as the words whose frequencies are more than 5\%.

\subsubsection{FMA}

The FMA dataset is a music dataset proposed by Defferrard $et~al.$ \cite{defferrard2016fma}. It extracts 140 audio features from 7,994 musics and is mainly used to classify music genres. This dataset also does not provide any graph.

\begin{table}[]
	\caption{Dataset information.}
	\label{tab:my-table}
	\begin{center}
		
	\begin{tabular}{lrrrr}
		\midrule
		Name     & Samples       & Features     & Graph & Train/Val/Test \\
		\midrule
		Wine     & 178           & 13           & No       & 10/20/158      \\
		Cacner   & 569           & 30           & No       & 10/20/539      \\
		Digits   & 1797 & 64           & No       & 50/100/1647    \\
		Citeseer & 3327          & 3703         & Yes        & 120/500/1000   \\
		Cora     & 2708          & 1433         & Yes        & 140/500/1000   \\
		20news   & 9607 & 236 & No       & 100/200/9307   \\
		FMA      & 7994          & 140          & No       & 160/320/7514  	\\
		\midrule
	\end{tabular}
\end{center}

\end{table}

\subsection{Baseline Methods }
When sample graph is not available, we compared our method with LDS, $k$NN-GAT, as well as two stage approaches methods mentioned in  \cite{franceschi2019learning} which first generate graphs and then use it to conduct node classification, including: 1) a sparse ER random graph (Sparse-GCN) \cite{erdHos1960evolution}; 2) a dense graph with equal edge probabilities (Dense-GCN); 3) a dense RBF kernel on the input features (RBF-GCN); and 4) a sparse k-nearest neighbor graph on the input features
(kNN-GCN). The methods in \cite{franceschi2019learning} also including some semi-supervised learning methods, including  label propagation (LP) \cite{zhu2003semi}, manifold regularization (ManiReg) \cite{belkin2006manifold}, and semi-supervised embedding (SemiEmb) \cite{weston2012deep}.
For these methods, we use the same results published in \cite{franceschi2019learning}.

When sample graph is available, we compared our method with following methods: 1) SemiSupervised GCN \cite{franceschi2019learning}, 2)  SemiEmb \cite{weston2012deep}, 3) DeepWalk \cite{perozzi2014deepwalk}, 4) ICA \cite{getoor2005link}, 5) Planetoid \cite{yang2016revisiting}, 6) GMNN \cite{qu2019gmnn}, 7) GRCN \cite{yu2019graph}, 8) GraphHeat \cite{xu2019graph},  9) GLCN \cite{jiang2019semi}, 10) DIAL-GNN \cite{chen2019deep}, and 11) Graphite \cite{grover2019graphite}.

\subsection{Implementation Details}
\label{details}

\begin{figure*}[htbp]

	\begin{center}
		\footnotesize

		\begin{tabular}{ccc}
			\centering
			\includegraphics[width=0.5\linewidth]{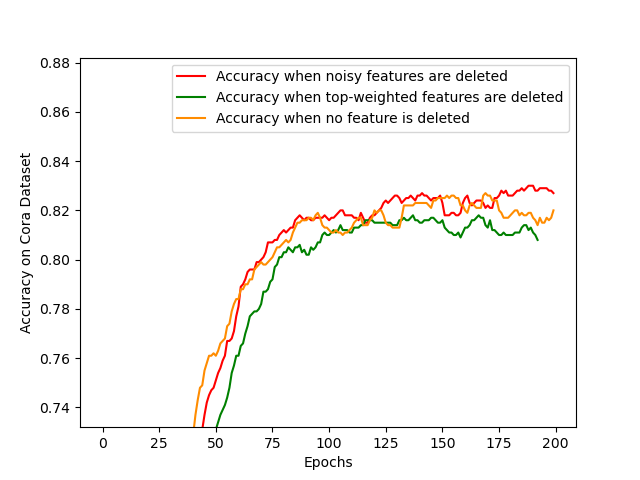}&  
			\includegraphics[width=0.5\linewidth]{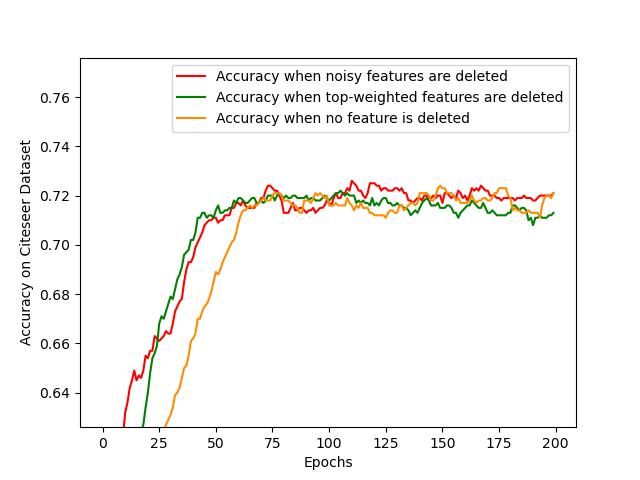}  \\
			(a) Accuracy on Cora Dataset &(b) Accuracy on Citeseer Dataset \\
			
		\end{tabular}
		
		\caption{Accuracy on Cora \textbf{(a)} and Citeseer \textbf{(b)} dataset after deleting noisy features, deleting the same number of top-weighted features and deleting no feature. After deleting noisy features, the accuracy does not change much. However, the accuracy decreases when the same number of top-weighted features are deleted.}
		\label{del_citeseer}
	\end{center}
	
\end{figure*}

For  data samples relation graph learning in \cref{sec311}, we select top-$k$ values of $\bR$ in \cref{adj_l1} and \cref{adj_2} to sparsify the learned graph.
We set an appropriate $k$ for each dataset using a validation dataset. For a dataset with a small feature dimension, we use a larger $k$ value (i.e, 90)
to use information from more nodes in the graph. When the number of features is large, we set $k$ to 10 or 20 to ensure the learned graph is sparse enough.

For feature updating in \cref{feature update section}, we use a two-layer generalized GCN model. The output dimension of the first layer is 32 and the activation function is the relu function. The output dimension of the second layer is the number of classes, and the softmax function is used as the activation function in the second layer. The dropout rate of the two GCN layers during the training process is set to 0.5. 



The values of the hyperparameters we used are summarized in \cref{tab:hyperparameters_table}.

\begin{table*}[]
	
	\caption{Summary of hyperparameters. Different datasets adopt the same values for regularization parameter, dropout rate and output dimension of generalized GCN layer 1.}
	\label{tab:hyperparameters_table}
	\begin{center}
		
	\begin{tabular}{cccccccc}
		\midrule
		Hyperparameter & Wine & Cancer & Digits & Citeseer & Cora & 20news & FMA \\
		\midrule
		Number of epochs              & 1000 & 160    & 700   & 48       & 200  & 1000    & 900 \\		
		k                  & 90   & 110     & 15     & 20       & 20   & 10     & 10  \\
		\midrule
		Regularization parameter   & \multicolumn{7}{c}{5.00E-04}                            \\
		Dropout rate		 & \multicolumn{7}{c}{0.5}                                 \\
		Output dimension of generalized GCN layer 1   & \multicolumn{7}{c}{32}                                 \\
		\midrule
	\end{tabular}
\end{center}

\end{table*}

\begin{figure*}[htbp]
	\begin{center}
		\footnotesize
		
		\begin{tabular}{ccc}
			\centering
			\includegraphics[width=0.4\linewidth]{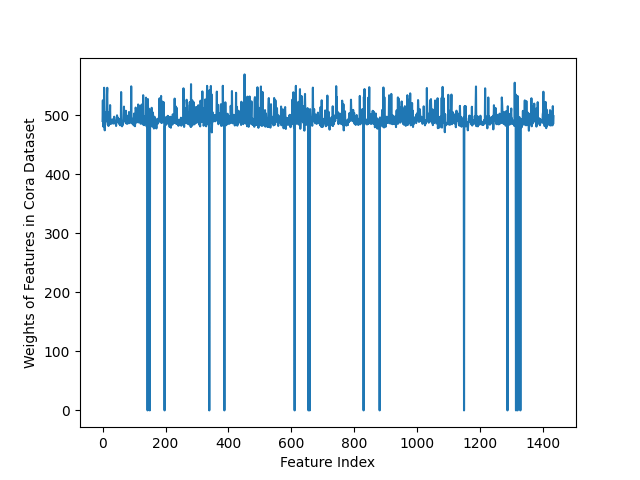}&
			\includegraphics[width=0.4\linewidth]{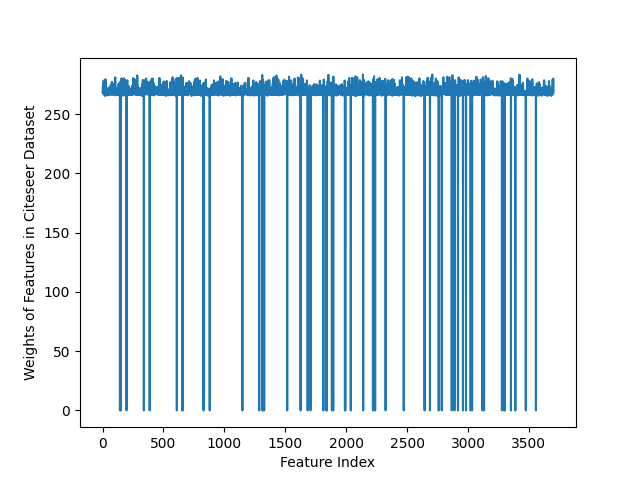}  \\
			(a) &(b) \\
			
		\end{tabular}
		\caption{ \textbf{(a)}: The weights of different features in Cora dataset.  Features corresponding to low weights are noisy features. \textbf{(b)}: The weights of different features in Citeseer dataset. Features corresponding to low weights are noisy features. }
		\label{fig:weights}
	\end{center}
\end{figure*}

\subsection{Sensitivity Analysis}

\begin{figure*}[htbp]
	
	\begin{center}
		\footnotesize
		
		\begin{tabular}{ccc}
			\centering
			\includegraphics[width=0.33\linewidth]{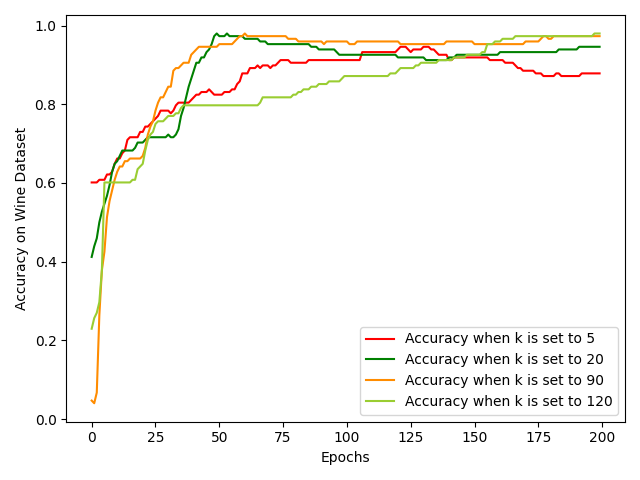}&
			\includegraphics[width=0.33\linewidth]{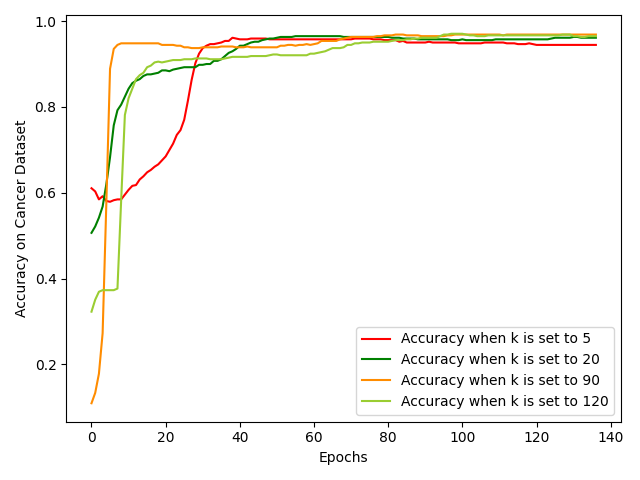}&
			\includegraphics[width=0.33\linewidth]{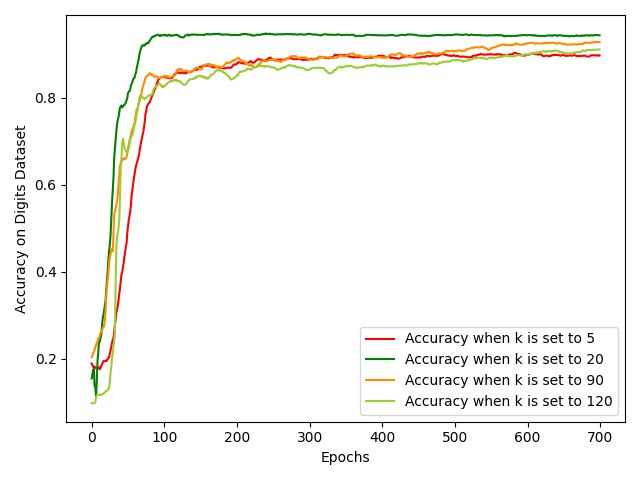} \\
			(a)Wine Dataset &(b)Cancer Dataset &(c)Digits Dataset\\
			
		\end{tabular}
		
		\begin{tabular}{ccc}
			\centering
			\includegraphics[width=0.33\linewidth]{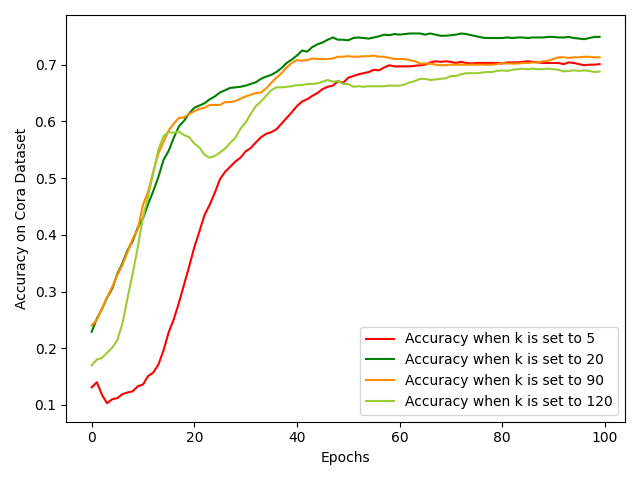}&
			\includegraphics[width=0.33\linewidth]{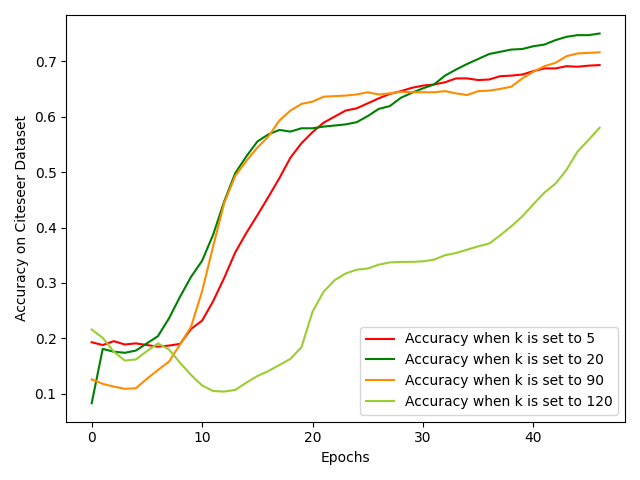}\\
			(d)Cora Dataset &(e) Citeseer Dataset\\
			
		\end{tabular}
		
		\caption{Impact of hyperparameter $k$ on Wine, Cancer, Digits, Cora, and Citeseer dataset.}
		\label{impacts_1}
		
	\end{center}
	
	\begin{center}
		\footnotesize
		\begin{tabular}{ccc}
			\centering
			\includegraphics[width=0.33\linewidth]{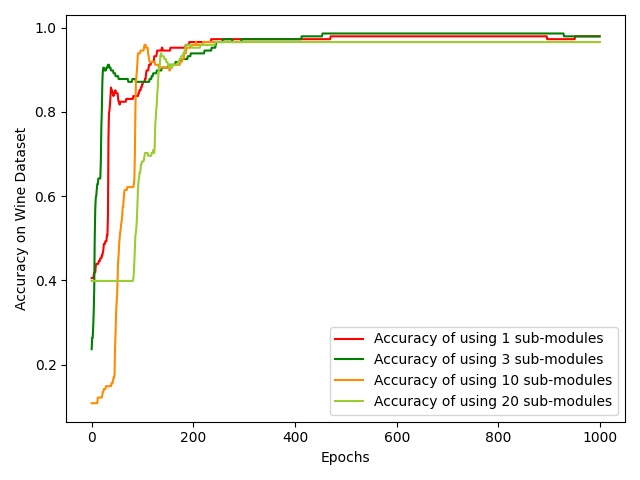}&
			\includegraphics[width=0.33\linewidth]{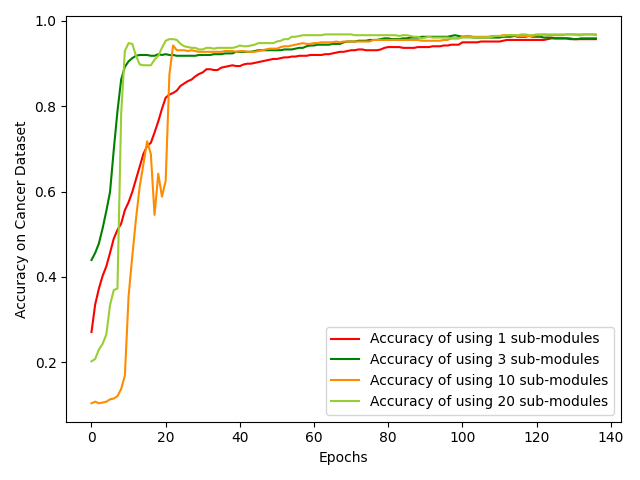}&
			\includegraphics[width=0.33\linewidth]{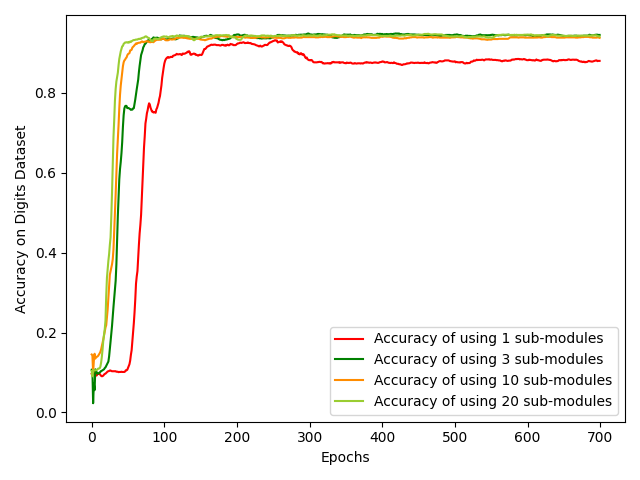} \\
			(a) Wine Dataset &(b) Cancer Dataset &(c) Digits Dataset\\
			
		\end{tabular}
		
		\begin{tabular}{ccc}
			\centering
			\includegraphics[width=0.33\linewidth]{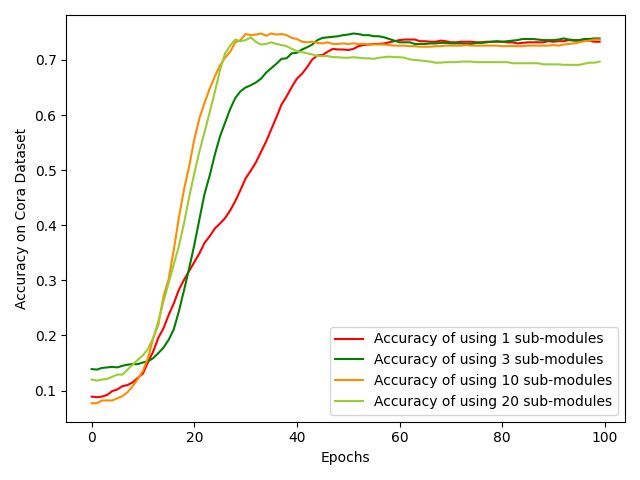}&
			\includegraphics[width=0.33\linewidth]{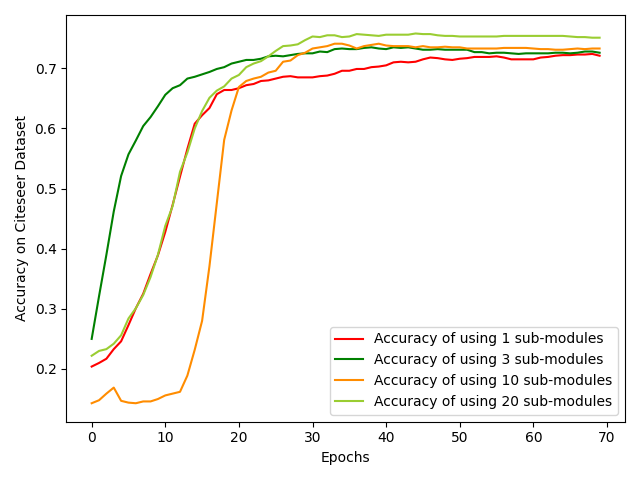}  \\
			(d) Cora Dataset &(e) Citeseer Dataset \\
			
		\end{tabular}
		
		\caption{Impact of hyperparameter $M$ on Wine, Cancer, Digits, Cora and Citeseer dataset.}
		\label{impacts}
	\end{center}
	
\end{figure*}

In this section, we will analyze the sensitivity of method to several hyperparameters. When analyzing each parameter, we fix other parameters and use the same settings as those in \cref{details}. 
\cref{impacts_1} and \cref{impacts} show how these hyperparameters impact the performance of our method.

\subsubsection{Impact of $k$} Different datasets have different appropriate values of $k$. For datasets with a large sample size, we set $k$ to a smaller value to make the learned graphs sparse enough.
\subsubsection{Impact of the number of sub-modules $M$} The results show larger number of sub-modules $M$ makes the algorithm converge faster. It can be observed that our method can achieve quite good results with a small number of sub-modules. We set $M$ to 3 in all the experiments.

\subsection{Results and Analysis}

In this part, we will present the experimental results and corresponding analysis from the perspective of the classification accuracy, the noise feature removal, and the noise relations removal. The feature and relations noise removal is equivalent to  identify the important features and relations, which can be used as the interpretable explanations of predictions defined in ~\cite{ying2019gnnexplainer}. 


\subsubsection{Node Classification Accuracy}
Experiments in this section aims at showing the performance of our network-of-graphs based model.
When external graph is not available, the accuracies of our algorithm and the baseline methods are shown in  \cref{tab:acc without graph}. We use bold fonts to highlight the best accuracies in the table. Our algorithm achieves higher node classification accuracies than other methods on all the datasets.

When an external graph is given in Cora and Citeseer dataset, the performance of our algorithm and the baseline methods are shown in \cref{tab:acc using graph}. In this case, our algorithm has the best performance on Citeseer dataset and is among one of the best methods on Cora dataset. 

\subsubsection{Key Feature Learning}
\label{sec45}
We do experiments on Cora and Citeseer datasets to show the effectiveness of our method on key feature learning. The results are shown in \cref{del_citeseer}. We delete the features corresponding to the small values in $\bw$ in \cref{selectfeature} and input the selected features into GCN\footnote{https://github.com/tkipf/gcn.} and generate better results than the original data, which shows that our method mainly remove the noisy features. In comparison, we also delete the same number of top-weighted features and input the data into GCN, and the accuracy decreases.
\cref{fig:weights} (a) shows the weights of different features on Cora dataset and \cref{fig:weights} (b) shows the weights of different features on Citeseer dataset. The noisy features of Cora and Citeseer datasets are corresponding to low weights.

\subsubsection{Key Graph Learning}

To show the ability of our method in learning key graph edges, we add a large number of noise edges to different datasets. For example, we added 7260 noise edges to Cora dataset, which have 10556 edges in total. The performance of our model is shown in \cref{tab:denoise}. We regard the original graph provided in the dataset as  the important graph edges and aim to evaluate how many noise edges can be removed using our method. We compute the difference among the noise edges added to the original graph and the remaining noise edges after the training process.

The weights of each noisy edges added to different datasets are randomly selected between 0 and 1. When the dataset contains a graph (such as Cora and Citeseer), we add the noise edges to the original graph and use it to train our model. When the dataset does not contain a graph (such as Wine, etc), we directly use the randomly generated noise edges to train our model.

The experimental results show that our algorithm can remove the added noise in most cases. As shown in \cref{tab:denoise},  our method can remove over 99\% added noises  on Digits, Cora, Citeseer, 20news and FMA datasets and 85\% noises on Cancer dataset. On Wine dataset, about 55\% noise edges are reserved, which may due to its small feature size.



\section{Conclusion}
\label{conclusion}
In this paper, we propose a graph neural network method called GL-GNN, which can deal with both the cases where there is a graph and there is no graph. Our method learns important features and important graph edges using multiple sub-modules, the results of these sub-modules are summarized by a learned sub-module network. Our method achieves better performance in node classification problems on multiple datasets than 24 baseline methods. Besides, our method also shows its capability in generating interpretable explanations of predictions. 


In our method, we only consider the data sample graph and can be further extended to learn feature graphs in the future.
\ifCLASSOPTIONcaptionsoff
  \newpage
\fi



\bibliographystyle{IEEEtran}
\bibliography{IEEEabrv,cite}
\end{document}

%% file: preamble.tex
\usepackage[T1]{fontenc}
\usepackage{amsmath,amssymb,amsfonts,mathrsfs,bm}
\usepackage{mathtools}
\usepackage{amsthm}
\usepackage[shortlabels]{enumitem}
\usepackage{graphicx}
\usepackage{epstopdf}
\usepackage{url}
\usepackage{colortbl}
\usepackage{multirow}
\usepackage{xcolor}
\usepackage[normalem]{ulem}
\usepackage{array}
\newcolumntype{L}[1]{>{\raggedright\let\newline\\\arraybackslash\hspace{0pt}}m{#1}}
\newcolumntype{C}[1]{>{\centering\let\newline\\\arraybackslash\hspace{0pt}}m{#1}}
\newcolumntype{R}[1]{>{\raggedleft\let\newline\\\arraybackslash\hspace{0pt}}m{#1}}

\makeatletter
\let\MYcaption\@makecaption
\makeatother
\usepackage[font=footnotesize]{subcaption}
\makeatletter
\let\@makecaption\MYcaption
\makeatother

\usepackage{xparse}



\usepackage{glossaries}
\newacronym{wrt}{w.r.t.}{with respect to}
\newacronym{RHS}{R.H.S.}{right-hand side}
\newacronym{LHS}{L.H.S.}{left-hand side}
\newacronym{iid}{i.i.d.}{independent and identically distributed}

\usepackage{float}

\usepackage[capitalize]{cleveref}
\crefname{equation}{}{}
\Crefname{equation}{}{}
\crefname{claim}{claim}{claims}
\crefname{step}{step}{steps}
\crefname{line}{line}{lines}
\crefname{dmath}{}{}
\crefname{dseries}{}{}
\crefname{dgroup}{}{}
\crefname{Theorem}{Theorem}{Theorems}
\crefname{Corollary}{Corollary}{Corollaries}
\crefname{Proposition}{Proposition}{Propositions}
\crefname{Lemma}{Lemma}{Lemmas}
\crefname{Definition}{Definition}{Definitions}
\crefname{Example}{Example}{Examples}
\crefname{Assumption}{Assumption}{Assumptions}
\crefname{Remark}{Remark}{Remarks}
\crefname{Rem}{Remark}{Remarks}
\crefname{remarks}{Remarks}{Remarks}
\crefname{Theorem_A}{Theorem}{Theorems}
\crefname{Corollary_A}{Corollary}{Corollaries}
\crefname{Proposition_A}{Proposition}{Propositions}
\crefname{Lemma_A}{Lemma}{Lemmas}
\crefname{Definition_A}{Definition}{Definitions}

\usepackage{algorithm,algorithmic}

\ifx\loadbreqn\undefined
	\relax
\else
	\usepackage{breqn} 
\fi

\ifx\renewtheorem\undefined
\ifx\useTheoremCounter\undefined
\newtheorem{Theorem}{Theorem}
\newtheorem{Corollary}{Corollary}
\newtheorem{Proposition}{Proposition}

\else

\fi

\fi

\theoremstyle{remark}

\theoremstyle{plain}

\newcommand{\calE}{\mathcal{E}}

\newcommand{\calV}{\mathcal{V}}
\newcommand{\calW}{\mathcal{W}}

\newcommand{\ba}{\boldsymbol{a}}
\newcommand{\bA}{\boldsymbol{A}}
\newcommand{\bb}{\boldsymbol{b}}

\newcommand{\bD}{\boldsymbol{D}}

\newcommand{\bg}{\boldsymbol{g}}

\newcommand{\bH}{\boldsymbol{H}}

\newcommand{\bI}{\boldsymbol{I}}

\newcommand{\bM}{\boldsymbol{M}}

\newcommand{\bN}{\boldsymbol{N}}

\newcommand{\bP}{\boldsymbol{P}}

\newcommand{\bR}{\boldsymbol{R}}
\newcommand{\bs}{\boldsymbol{s}}

\newcommand{\bV}{\boldsymbol{V}}
\newcommand{\bw}{\boldsymbol{w}}
\newcommand{\bW}{\boldsymbol{W}}
\newcommand{\bx}{\boldsymbol{x}}
\newcommand{\bX}{\boldsymbol{X}}
\newcommand{\by}{\boldsymbol{y}}
\newcommand{\bY}{\boldsymbol{Y}}

\newcommand{\bZ}{\boldsymbol{Z}}

\newcommand{\balpha}{\bm{\alpha}}

\newcommand{\tbx}{\widetilde{\bx}}
\newcommand{\tbX}{\widetilde{\bX}}

\newcommand{\qednew}{\nobreak \ifvmode \relax \else
      \ifdim\lastskip<1.5em \hskip-\lastskip
      \hskip1.5em plus0em minus0.5em \fi \nobreak
      \vrule height0.75em width0.5em depth0.25em\fi}

\newcommand{\cond}[2]{\left. {#1}\, \middle| \, {#2} \right.}

\DeclareDocumentCommand \P { g d() g } {%
	\IfNoValueTF {#3} 
	{%
		\IfNoValueTF {#1} 
		{%
			\IfNoValueTF {#2}
			{%
				\mathbb{P}%
			}%
			{%
				\mathbb{P}\left({#2}\right)%
			}%
		}%
		{%
			\IfNoValueTF {#2}
			{%
				\mathbb{P}_{#1}%
			}%
			{%
				\mathbb{P}_{#1}\left({#2}\right)%
			}%
		}%
	}%
	{%
		\IfNoValueTF {#1} 
		{%
			\mathbb{P}\left(\cond{#2}{#3}\right)%
		}%
		{%
			\mathbb{P}_{#1}\left(\cond{#2}{#3}\right)%
		}%
	}%
}

\DeclareDocumentCommand \E { g o g } {%
	\IfNoValueTF {#3} 
	{%
		\IfNoValueTF {#1} 
		{%
			\IfNoValueTF {#2}
			{%
				\mathbb{E}%
			}%
			{%
				\mathbb{E}\left[{#2}\right]%
			}%
		}%
		{%
			\IfNoValueTF {#2}
			{%
				\mathbb{E}_{#1}%
			}%
			{%
				\mathbb{E}_{#1}\left[{#2}\right]%
			}%
		}%
	}%
	{%
		\IfNoValueTF {#1} 
		{%
			\mathbb{E}\left[\cond{#2}{#3}\right]%
		}%
		{%
			\mathbb{E}_{#1}\left[\cond{#2}{#3}\right]%
		}%
	}%
}

\definecolor{gray90}{gray}{0.9}

\ifx\nohighlights\undefined

	\newcommand{\msout}[1]{\text{\color{green} \sout{\ensuremath{#1}}}}
	\newcommand{\del}[1]{{\color{green}\ifmmode \msout{#1}\else\sout{#1}\fi}}
\else

	\newcommand{\msout}[1]{#1}
	\newcommand{\del}[1]{#1}
\fi

\newcommand{\hide}[1]{}

\renewcommand{\figurename}{Fig.}

\graphicspath{{./Figures/}} 
\ifx\diagnoselabel\undefined
	\relax
\else
	\makeatletter
	 \def\@testdef #1#2#3{%
		 \def\reserved@a{#3}\expandafter \ifx \csname #1@#2\endcsname
		\reserved@a  \else
	 \typeout{^^Jlabel #2 changed:^^J%
	 \meaning\reserved@a^^J%
	 \expandafter\meaning\csname #1@#2\endcsname^^J}%
	 \@tempswatrue \fi}
	\makeatother
\fi
